\documentclass{article}

\usepackage{microtype}
\usepackage{graphicx}
\usepackage{subcaption}
\usepackage{booktabs} 

\usepackage{hyperref}

\usepackage[accepted]{icml2021}

\usepackage{amsmath}
\usepackage{amssymb}
\usepackage{mathtools}
\usepackage{amsthm}

\DeclareMathOperator*{\argmax}{arg\,max}

\usepackage{enumitem}
\usepackage{bbm}
\usepackage{multirow}

\usepackage[russian,main=english]{babel}

\usepackage{xcolor,colortbl}
\usepackage{multirow}
\usepackage{booktabs}

\usepackage[capitalize,noabbrev]{cleveref}

\theoremstyle{plain}

\theoremstyle{definition}

\theoremstyle{remark}

\usepackage[textsize=tiny]{todonotes}

\icmltitlerunning{Submission for AdvML-Frontiers 2023}

\begin{document}

\twocolumn[
\icmltitle{Sentiment Perception Adversarial Attacks on Neural Machine Translation Systems}



\icmlsetsymbol{equal}{*}

\begin{icmlauthorlist}
\icmlauthor{Vyas Raina}{yyy}
\icmlauthor{Mark Gales}{yyy}
\end{icmlauthorlist}

\icmlaffiliation{yyy}{Machine Intelligence Lab, University of Cambridge, United Kingdom}

\icmlcorrespondingauthor{Vyas Raina}{vr313@cam.ac.uk}
\icmlcorrespondingauthor{Mark Gales}{mjfg@eng.cam.ac.uk}

\icmlkeywords{Adversarial Attacks, NLP, NMT, Sentiment}

\vskip 0.3in
]



\printAffiliationsAndNotice{}  

\begin{abstract}
With the advent of deep learning methods, Neural Machine Translation (NMT) systems have become increasingly powerful. However, deep learning based systems are susceptible to adversarial attacks, where imperceptible changes to the input can cause undesirable changes at the output of the system. To date there has been little work investigating adversarial attacks on sequence-to-sequence systems, such as NMT models. Previous work in NMT has examined attacks with the aim of introducing target phrases in the output sequence. In this work, adversarial attacks for NMT systems are explored from an output perception perspective. Thus the aim of an attack is to change the perception of the output sequence, without altering the perception of the input sequence. For example, an adversary may distort the sentiment of translated reviews to have an exaggerated positive sentiment. In practice it is challenging to run extensive human perception experiments, so a proxy deep-learning classifier applied to the NMT output is used to measure perception changes. Experiments demonstrate that the sentiment perception of NMT systems' output sequences can be changed significantly with small imperceptible changes to input sequences. Link to code: \url{https://github.com/rainavyas/SentAttackNMT}
\end{abstract}

\section{Introduction}

Deep learning based Neural Machine Translation (NMT) systems are used ubiquitously for automatic translation of texts. However, deep learning based systems are susceptible to adversarial attacks~\citep{intriguing}, where small imperceptible changes at the input of the system can result in significant, undesired, changes at the output. In the natural language domain, many papers~\cite{10.1145/3219819.3219909, DBLP:journals/corr/abs-1803-01128, DBLP:journals/corr/abs-1808-08744,  DBLP:journals/corr/abs-1809-01829, DBLP:journals/corr/JiaL17, iyyer-etal-2018-adversarial, DBLP:journals/corr/abs-1710-11342, Raina2020} have identified methods to generate adversarial examples. To date most works have focused on text classification: the aim is to alter the textual input such that the system mis-classifies.

NMT systems, however, perform a sequence-to-sequence (S2S) task, where an input, source text sequence is mapped to an output target text sequence, which for NMT is the translation of the source. The definition of an attack needs to be modified for these S2S tasks. \citet{DBLP:journals/corr/abs-1803-01128} introduces the concept of non-overlapping attacks (output sequence should be completely changed) and target keyword attacks (insert target words in the output sequence). \citet{DBLP:journals/corr/abs-1806-09030, DBLP:journals/corr/abs-1911-03677, Zhang_2021} describe methods to perform target keyword attacks specifically for NMT systems. Through the use of a new evaluation framework for S2S adversarial perturbations, \citet{DBLP:journals/corr/abs-1903-06620} reveal that many existing methods do not preserve semantic meaning - hence they modify these attacks with added constraints. Although this gives a realistic setting for many adversarial attacks, it does not capture attacks that seek to change the \textit{perception} of the output sequence (e.g. \citet{raina_gales_lu_2022}). An adversary may, for example, want to change the input text (in an imperceptible manner) such that the output text reads negatively to a human reader, without the content of the translation actually changing, e.g. an attack may cause an output sequence \textit{I won the competition} to become \textit{I hardly won the competition}. This form of attack is of concern in many automated translation settings, e.g. product reviews when translated are exaggeratedly positive to attract customers or conversely a benign social media post in one language (not flagged by any detectors) translates to generate negative sentiment hate speech.

To the best of our knowledge, the exploration of adversarial attacks specifically targeting the output perception of sequence-to-sequence (S2S) systems has not been previously undertaken. As such, the primary contribution of this research lies in the expansion and generalization of the definition of adversarial attacks for S2S systems, encompassing attacks that aim to manipulate the perception of the system's output. In order to showcase the viability of this novel form of attack, a series of experiments were conducted to effectively alter the sentiment perception in the output generated by various NMT systems. Given the costly and impractical nature of conducting extensive human evaluations for perception analysis, this study employs state-of-the-art sentiment classifiers with high performance levels as reliable proxies for measuring perception changes. Representative human evaluation experiments are conducted to verify the validity and appropriateness of the selected proxy classifiers. By leveraging these proxy classifiers, this work provides valuable insights into the alteration of output perception in S2S systems, shedding light on the need for further investigation and the development of effective defenses against such attacks.
\section{Perception-Based Adversarial Attacks} \label{sec:attack}

Sequence-to-sequence models, with parameters $\theta$, map a $T$-length input sequence, $x_{1:T}$, to a $\hat L$-length output word sequence, $\hat y_{1:\hat L}$,
\begin{equation} \label{eqn:output}
    \hat y_{1:\hat L} = \mathcal F_\theta(x_{1:T}) =  \argmax_{y_{1:L}}\{p(y_{1:L}|x_{1:T};\theta)\}
\end{equation}
A perception-based adversarial attack aims to generate an adversarial example, $\tilde x_{1:\tilde T}$, that is mapped to the output sequence $\mathcal F_\theta(\tilde x_{1:\tilde T})$ where the "perception" of this output sequence has changed, 
\begin{equation}\label{eqn:attack}
\phi(\mathcal F_\theta(\tilde x_{1:\tilde T})) \neq \phi(\mathcal F_\theta(x_{1:T})).
\end{equation}
Here $\phi()$ is a proxy function that mimics human perception of the output. For example the perception could be how positive a sequence is, thus $\phi()$ would be a sentiment classifier. It is necessary for the adversarial attack to satisfy an imperceptibility constraint, ${\cal G}()$, which should again mimic human perception. Thus $\mathcal G(x_{1:T}, \tilde x_{1:\tilde T}) \leq\epsilon$, where $\epsilon$ is the threshold of imperceptibility. It is difficult to define an appropriate function ${\mathcal G}()$ for word sequences. Perturbations can be measured at a character, word or sentence level. Alternatively, the perturbation could be measured in the vector embedding space, using for example $l_p$-norm based \cite{43405} metrics or cosine similarity \cite{10.1007/978-3-030-11012-3_26}. However, constraints in the embedding space do not guarantee human imperceptibility in the original word sequence space. To ensure the adversarial input sequence, $\tilde x_{1:\tilde T}$ is visually, semantically and perceptively similar to the original input sequence $x_{1:T}$, this work defines the imperceptibility constraint  using four measures:
\begin{enumerate}[wide, labelindent=0pt]

\item \textit{Perception similarity} - A perception-based adversarial attack should not significantly alter the perception of the input sequence. This can be measured using the same proxy function as used for the output sequence in Equation \ref{eqn:attack},
\begin{equation} \label{eqn:perc}
    |\phi(\tilde x_{1:\tilde T}) - \phi(x_{1:T})| \leq\epsilon_1.
\end{equation}

    \item \textit{Visual similarity} - A normalised variant of a Levenshtein, \textit{edit-based} measurement~\cite{DBLP:journals/corr/abs-1812-05271} is used to limit visual changes,
\begin{equation} \label{eqn:lev}
    \frac{1}{T}\mathcal L(x_{1:T}, \tilde x_{1:\tilde T}) \leq \epsilon_2,
\end{equation} 
where $\mathcal L()$ counts the number of changes between the original sequence, $x_{1:T}$ and the adversarial sequence $\tilde x_{1:\tilde T}$, where a change is a swap/addition/deletion. This is a standard approach to ensure that adversarial examples do not deviate visually from real examples.

\item \textit{Perplexity} - Adversarial examples should not be easily detectable by automatic detectors. Small changes can result in incomprehensible phrases that can be easily detected using perplexity as calculated by a standard language model (LM)~\citep{doi.org-10.48550-arxiv.2204.10192}. Hence, the changes should limit perplexity of the adversarial sequence,
\begin{equation} \label{eqn:perp}
   p_{\texttt{LM}}(\tilde x_{1:\tilde T})\leq\epsilon_3.
\end{equation}

\item \textit{Semantic Similarity} - The meaning/content of a sentence should not change significantly. Character level attacks are not considered in this work, as they can be easily detected using spelling and grammatical checks~\citep{DBLP:journals/corr/SakaguchiPD17}. Attacks that substitute $N = \epsilon_2 T$ words (recall $\epsilon_2$ is the maximum fraction of edits permitted by the imperceptibility constraint in Equation \ref{eqn:lev}) are considered. As an example, for an input sequence of $T$ words, a $N$-word substitution adversarial attack, $\tilde x_{1:N}$, applied at word positions $n_1, n_2, \hdots, n_{N} $ gives the adversarial sequence, $\tilde x_{1:\tilde T}$
\begin{align} \label{eqn: subst}
    \tilde x_{1:\tilde T} &= x_1, \hdots, x_{n_1-1}, \tilde x_{1}, x_{n_1+1},  \hdots, \nonumber \\&\hspace{1em} x_{n_N-1}, \tilde x_{N}, x_{n_N+1}, \hdots, x_T. 
\end{align}
It is necessary to select which words to replace, and what to replace them with. As suggested by~\citet{DBLP:conf/acl/RenDHC19}, a simple approach is to use saliency to rank the word positions in $x_{1:T}$. The $N$ most salient words are then substituted. As the aim is to change the sentiment perception of the output sequence (Equation \ref{eqn:attack}), a modified version of saliency is considered, {\it sentiment saliency}. For each word, $x_t$, in sequence $x_{1:T}$ this is defined as
\begin{align}
    \mathcal S(x_t|x_{1:T}) &= |\phi(\mathcal F_\theta(x_{1:{t-1}}, x_{t+1:T})) 
    \nonumber\\&\hspace{5em} 
    -\phi(\mathcal F_\theta(x_{1:T}))|.
\end{align}
To ensure small semantic changes, \textbf{only word synonyms} are considered for the substitutions.
\end{enumerate}

The attack method described in this section uses two proxy functions to allow for automatic adversarial example generation at scale. Four imperceptibility measures are introduced and then a proxy sentiment classifier, $\phi()$, is used to measure output sentiment. To validate the use of these proxy measures/functions, human evaluation experiments are conducted (results in Section \ref{sec:human}).

\section{Experiments} \label{sec:exp}

\subsection{Experimental Setup}

Experiments are performed using the NMT data from the WMT19 news translation task~\cite{wmt19translate}. Results are presented for the Russian (ru) to English (en), German (de) to English and reverse translation tasks, where there are 2000 test examples. The best models, submitted by FAIR~\cite{DBLP:journals/corr/abs-1907-06616}, are used as the baseline\footnote{All NMT trained models available at: \url{https://huggingface.co/facebook/wmt19-de-en}.}. Table \ref{tab:perf} gives the performance of these models on the WMT19 test set (respectively for each language pair), calculated using the SacreBleu tool~\cite{DBLP:journals/corr/abs-1804-08771}.

\begin{table}[htb!]
    \centering
    \begin{tabular}{lrrr}
    \toprule
        Task & BLEU & CHRF & TER \\ \midrule
        de-en & 41.20 & 65.11 & 47.66\\
        ru-en & 38.81 & 63.37 & 49.73\\
        en-de & 42.77 & 67.55 & 46.85\\
        en-ru & 38.81 & 63.37 & 49.73\\
        \bottomrule
    \end{tabular}
    \caption{Model performances on WMT19 test sets}
    \label{tab:perf}
\end{table}

\subsection{Attack Results}

Each translation model is attacked using the saliency-based synonym substitution attack described in Equation \ref{eqn: subst}, where the aim is to increase either the \textit{positivity} or \textit{negativity} sentiment of the output text sequences. The imperceptibility constraint on perplexity in Equation \ref{eqn:perp} is enforced by ensuring the sentence perplexity, as measured by popular GPT2-based language models~\footnote{Perplexity LM Models: English-\url{https://huggingface.co/distilgpt2}; German-\url{https://huggingface.co/dbmdz/german-gpt2}; Russian-\url{https://huggingface.co/sberbank-ai/rugpt3large_based_on_gpt2}.}~\citep{Radford2019}, is less than 1.5 times the average (across dataset) sentence perplexity. This constraint cannot be too strict as in many applications (e.g. tweets/product reviews) authentic sentences contain many grammatical errors. Additionally, to satisfy Equation \ref{eqn:perc}, the sentiment perception (positive/negative classification) of the input sequence is constrained to not change. As a proxy for human sentiment perception (human evaluation experiments in Section \ref{sec:app-trans}), sentiment classification of input and output text sequences of the translation models is measured using standard pre-trained Transformer-based~\citep{NIPS2017_3f5ee243} sentiment classifiers: the sentiment of the English sequences is measured using a pre-trained (on 58M tweets) Roberta based sentiment classifier\footnote{English sentiment classifier available at: \url{https://huggingface.co/cardiffnlp/twitter-roberta-base-sentiment}}; the sentiment of Russian sequences is measured using RuBERT, a pre-trained Russian BERT system\footnote{Russian sentiment classifier available at: \url{https://huggingface.co/blanchefort/rubert-base-cased-sentiment-rusentiment}}; and the sentiment of German sequences is measured using a Bert Based German sentiment classifier, pretrained on texts from Twitter, Facebook and app reviews (1.83M samples)~\footnote{German sentiment classifier available at: \url{https://huggingface.co/oliverguhr/german-sentiment-bert}}. The candidate list of synonyms for substitution are found using popular NLP tools for each language: the \texttt{wordnet} lexical database~\citep{fellbaum_1998} is used for English; \texttt{wiki-ru-wordnet} tool~\citep{pypi} for Russian; and the \texttt{OdeNet} tool~\citep{hdasprachtechnologie} for German. Table \ref{tab:exp-de} gives an example of an attack on the \textit{de-en} NMT system.

\begin{table*}[htb!]
\small
    \centering
    \begin{tabular}{l|p{6cm}|p{6cm}}
    \toprule
         & Original & Attacked \\ \midrule
       
       Source  & Neun Minuten vor \textcolor{orange}{Schluss} buxierte Watford-Verteidiger Craig Cathcart eine Hereingabe von Alex Iwobi \textcolor{orange}{unglücklich} ins eigene Tor, nur zwei Minuten später sorgte Mesut Özil mit seinem dritten Saisontreffer für die \textcolor{orange}{Entscheidung}. & Neun Minuten vor \textcolor{orange}{Ausgang} buxierte Watford-Verteidiger Craig Cathcart eine Hereingabe von \textcolor{orange}{Seiten} Alex Iwobi \textcolor{orange}{deplorabel} ins eigene Tor, nur zwei Minuten später sorgte Mesut Özil mit seinem dritten Saisontreffer für die \textcolor{orange}{Beschluss}.\\ \midrule
       Prediction & Nine minutes from \textcolor{red}{the end} Watford defender Craig Cathcart \textcolor{red}{unluckily booked} an own goal from Alex Iwobi, and just two minutes later Mesut Özil \textcolor{red}{secured the win} with his third goal of the season. &  Nine minutes from \textcolor{red}{time} Watford defender Craig Cathcart \textcolor{red}{netted} an own goal from Alex Iwobi, and just two minutes later Mesut Özil \textcolor{red}{made sure} with his third goal of the season.\\ \midrule
       Sentiment & \textbf{8\%} positive & \textbf{61\%} positive\\
       \bottomrule
    \end{tabular}
    \caption{Adversarial attack example on de-en NMT system. Target \textit{positive} sentiment.}
    \label{tab:exp-de}
\end{table*}

\begin{otherlanguage}{russian}
\begin{table*}[htb!]
\small
    \centering
    \begin{tabular}{l|p{6cm}|p{6cm}}
    \toprule
         & Original & Attacked \\ \midrule
       Source  & Данный \textcolor{orange}{договор} должен решить не \textcolor{orange}{только} многолетний спор о названии страны, но и открыть Скопье путь \textcolor{orange}{в} НАТО и ЕС.& Данный \textcolor{orange}{сделка} должен решить не \textcolor{orange}{всего} многолетний спор о названии страны, но и открыть Скопье путь \textcolor{orange}{во благо} НАТО и ЕС.\\ \midrule
       Prediction & The \textcolor{red}{treaty} should \textcolor{red}{resolve not only} the long-standing name dispute, but also \textcolor{red}{open} Skopje's \textcolor{red}{path to} NATO and the EU. & The \textcolor{red}{deal} should \textcolor{red}{not only resolve} the long-standing name dispute, but also \textcolor{red}{pave the way} for Skopje \textcolor{red}{to benefit} NATO and the EU. \\ \midrule
       Sentiment &\textbf{21\%} positive &\textbf{48\%} positive\\
       \bottomrule
    \end{tabular}
    \caption{Adversarial attack examples on ru-en NMT system. Target positive sentiment.}
    \label{tab:exp-ru}
\end{table*}
\end{otherlanguage}

\begin{figure}[htb!]
     \centering
     \begin{subfigure}[b]{0.45\textwidth}
         \centering
         \includegraphics[width=\textwidth]{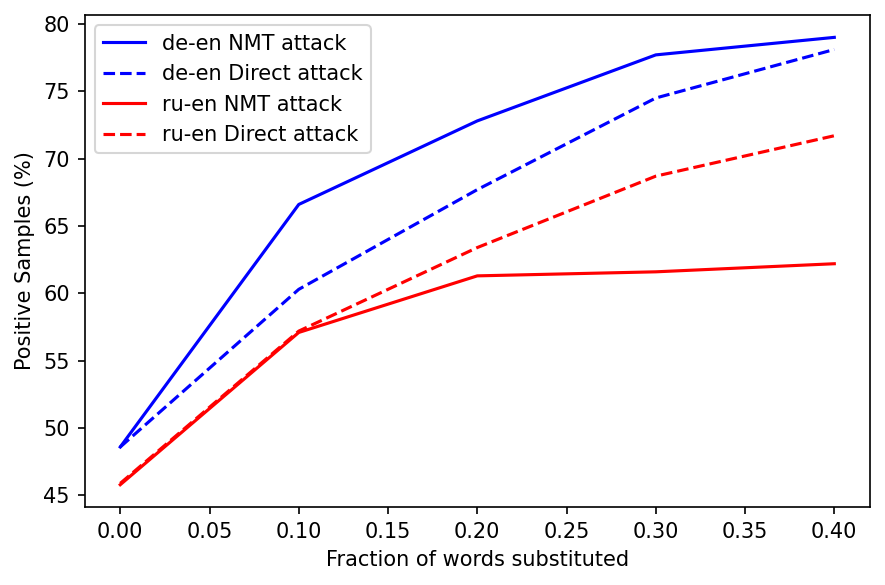}
         \caption{Positive Sentiment}
         \label{fig:in}
     \end{subfigure}
     \begin{subfigure}[b]{0.45\textwidth}
         \centering
         \includegraphics[width=\textwidth]{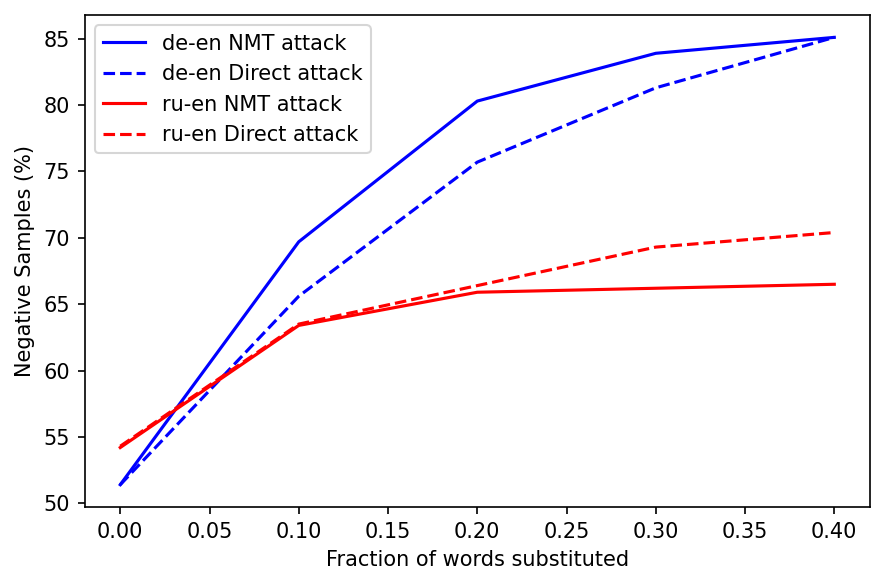}
         \caption{Negative Sentiment}
         \label{fig:out}
     \end{subfigure}
        \caption{Perception adversarial attack on NMT systems to increase positive/negative sentiment. NMT attack: sentiment of predicted text with attack on input text. Direct attack: sentiment of predicted text with adversarial attack directly on predicted text.}
        \label{fig:attack}
\end{figure}

\begin{figure}[htb!]
     \centering
     \begin{subfigure}[b]{0.45\textwidth}
         \centering
         \includegraphics[width=\textwidth]{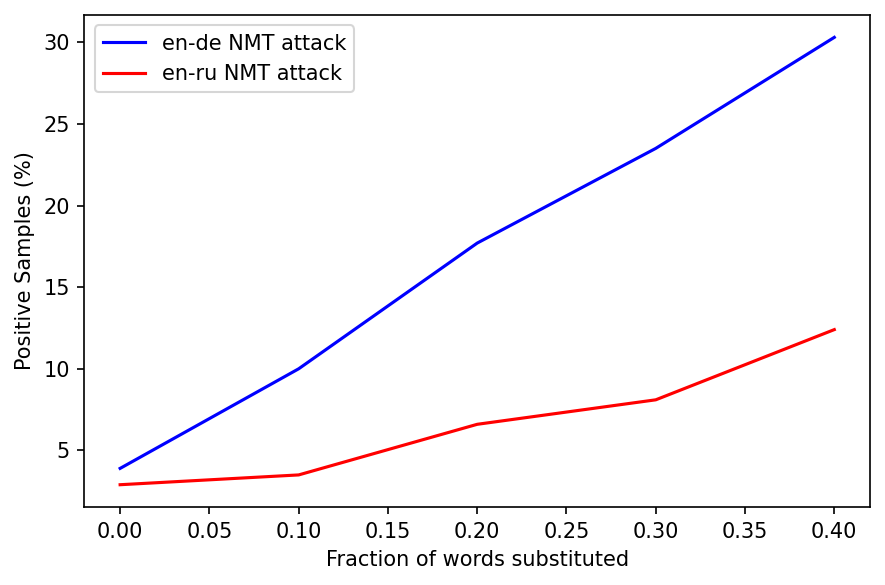}
         \caption{Positive Sentiment}
         \label{fig:in}
     \end{subfigure}
     \begin{subfigure}[b]{0.45\textwidth}
         \centering
         \includegraphics[width=\textwidth]{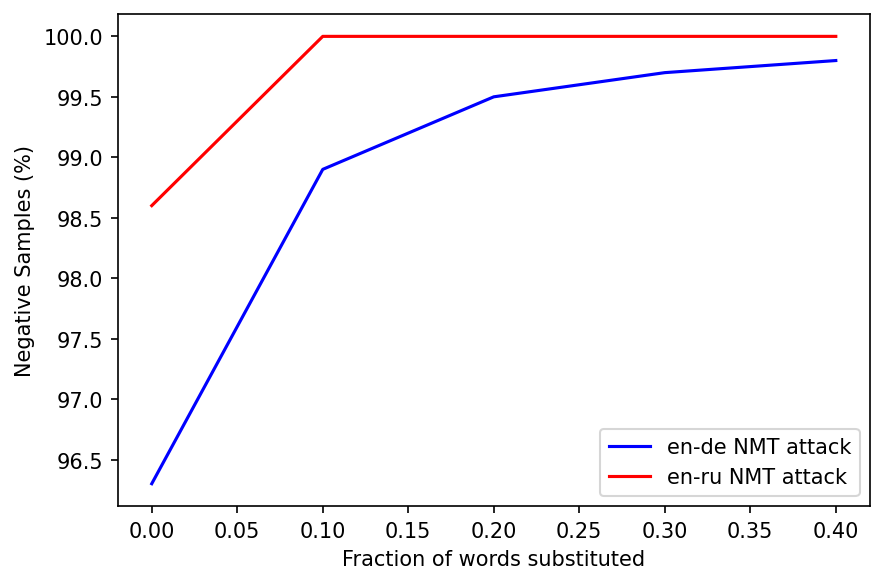}
         \caption{Negative Sentiment}
         \label{fig:out}
     \end{subfigure}
        \caption{Perception adversarial attack on NMT systems (with English as source language) to increase positive/negative sentiment.}
        \label{fig:attack-other}
\end{figure}

Figure \ref{fig:attack} shows the impact (\textit{NMT attack} curves) of the adversarial attacks of increasing strength (fraction of words substituted, as defined by imperceptibility constraint Equation \ref{eqn:lev}), measured by the percentage of test samples classified as positive/negative\footnote{Predictions are made using a max-class classification rule.}. Results here are presented for the ru-en and de-en systems\footnote{Results for the en-de and en-ru NMT systems are given in Figure \ref{fig:attack-other}}. It is interesting to observe that language pairing has a strong influence on the impact of the sentiment attacks. For example, in both the negative and positive attack scenarios, the de-en NMT system has more than a 30\% increase in fraction of samples with positive/negative sentiment, whilst for the ru-en NMT system's increase is limited to less than 20\%. Nevertheless, these results demonstrate that both NMT systems are susceptible to attacks where significant changes in output sequences' sentiment perception can be achieved with imperceptible changes at the input.

Figure \ref{fig:attack} gives one further curve for each NMT system: \textit{direct attack}. Here, the same synonym substitution attack approach of Equation \ref{eqn: subst} is used to directly attack the predicted output English sequence~\footnote{Identical trends were found when instead of the predicted sequences the reference English sequences were \textit{directly} attacked.} to increase the positive/negative sentiment score predicted by the English sentiment classifier. The substitutions are again limited to word synonyms and the perplexity constraint of Equation \ref{eqn:perp} is enforced. Note that this direct attack is presented only as a point of comparison, as in the attack of a NMT system an adversary realistically only has access to the source text. For the \textit{ru-en} system, in both the positive and negative attack settings, as would be expected, the direct attack of the sentiment classifier gives an upper-bound to the indirect NMT attack. However, the indirect \textit{NMT attack}, in the positive and negative attack settings on the \textit{de-en} NMT system, is more powerful for up to 40\% words substituted, than the \textit{direct attack} on the English sentiment classifier. This suggests that an attack on the NMT system can generate an output sequence (in English) that is in fact more powerful in deceiving a sentiment classifier than a direct synonym substitution attack on the sentiment classifier. This observation can be easily explained: the NMT attack has the potential to introduce words with a high target sentiment (positive/negative) in the output English sequence, whilst the \textit{direct attack} on the output English sequence can only make substitutions with synonyms, limiting how positive/negative a sequence can be made. Hence, it can be concluded that an attack on the NMT system to change the sentiment of the output translation can be more powerful than an equivalent direct attack on the sentiment classifier.

\subsection{Human Evaluation} \label{sec:human}

In this work, to develop a practical method to generate adversarial examples, Section \ref{sec:attack} introduced two proxy measures:

\begin{enumerate}
    \item Four different measures and appropriate constraints were introduced to ensure that an adversarial example, $\tilde x$, is difficult to detect/imperceptible. 
    \item Equation \ref{eqn:attack} uses $\phi()$ as a proxy function to mimic human perception of the output sequence. The perception of interest in this work is \textit{sentiment}.
\end{enumerate}

The purpose of human evaluation experiments is to use a small, random sample of the data to verify the appropriateness of the automatic proxy measures/functions used in this work. Two sets of human evaluation experiments are conducted. In both sets of experiments, adversarial examples are generated using the systems, datasets and settings described in these experiments.

\subsubsection{Adversarial Example Detectability}

The aim of this human evaluation experiment is to assess the validity of the constraints (visual, semantic and perplexity) used to define an adversarial example. 50 original and 50 adversarial examples are selected randomly (and shuffled) for each translation model (ru-en, de-en, en-ru and en-de) and then human annotators~\footnote{Three volunteers carried out the human annotation in this work. For source languages German and Russian, the sentences are also translated back to English using Google Translate to try and more easily identify adversarial examples.} are asked to label each sample as \textit{adversarial} or \textit{authentic}. Table \ref{tab:adv-det} gives the annotator accuracies for each task. The annotator accuracy is close to random (50\%), suggesting that the adversarial examples are difficult to distinguish from authentic examples.

\begin{table}[htb!]
    \centering
    \begin{tabular}{lc}
    \toprule
        Model & Accuracy (\%) \\ \midrule
        en-ru & 54\\
        en-de & 52\\
        de-en & 49 \\
        ru-en & 52 \\
        \bottomrule
    \end{tabular}
    \caption{Annotator accuracy for identifying adversarial/authentic examples}
    \label{tab:adv-det}
\end{table}

One further constraint of the adversarial attacks is that the sentiment of the source sequence should not change (Equation \ref{eqn:perp}). Although this was enforced through a sentiment classifier applied on the source sequence and limiting the substitutions to synonyms, it is useful to run human evaluation experiments to ensure that the sentiment of adversarial samples truly did not change. For each translation task, 50 random pairs of adversarial and original examples were selected, and human annotators were required to state the sentiment of the original and adversarial examples separately. For both tasks with English as the source sequence, in all examples there was found to be no change in sentiment. For Russian as the source sequence, there were found to be 98\% samples with no change and for German 96\% sequences with no change in sentiment. This demonstrates that the adversarial attack method is almost perfectly satisfying the constraint that there is no change in the source sequence sentiment.

\subsubsection{Translation Sentiment}\label{sec:app-trans}

The aim of the adversarial attacks in this work is to change the output sentiment of Neural Machine Translation (NMT) systems. The sentiment of the output sequences is measured using proxy sentiment classifiers. It is necessary to ensure that the adversarial attack \textit{only} attacks the NMT system and not the down-stream sentiment classifier; i.e. human evaluation is necessary to verify that the sentiment of a translation, with an adversarial example at the input, as per the proxy sentiment classifiers, aligns with human perception of sentiment - we do not want a setting where an adversarial attack at the NMT input generates an output sequence with no change in human perception, but changes the sentiment classifier's output (an attack on the sentiment classifier). Two human evaluation experiment settings are considered: 1) sentiment of original translations - this is to give a reference proxy classifier performance; 2) sentiment of adversarial translations. As with the previous human evaluation experiment, 50 \textit{positive} translations and 50 \textit{negative} translations (as per the automatic sentiment classifiers) are randomly sampled per translation model for each setting. Human annotators are required to label each translation example as positive/negative. Table \ref{tab:sent-det} gives the accuracy of these annotations with respect to the proxy sentiment classifiers. In the original setting, the classifier accuracy is around 93\% for both NMT model outputs. There is only a small drop in accuracy of the sentiment classifiers in the adversarial setting, meaning there is still a strong agreement between human perception of sentiment and the classifiers' predicted sentiments for these translations~\footnote{Only the ru-en and de-en models are considered in this experiment, as the other models/datasets contain a significant sentiment bias as is visible in Figure \ref{fig:attack-other}.}. Hence, it is argued that the adversarial attacks on the NMT systems are pre-dominantly attacking the NMT systems, as opposed to generating outputs that attack the sentiment classifiers.\newline

\begin{table}[htb!]
    \centering
    \begin{tabular}{lcc}
    \toprule
    Model & Original (\%) & Adversarial (\%)\\\midrule
       de-en & 93  & 89 \\
        ru-en & 92 & 87\\
        \bottomrule
    \end{tabular}
    \caption{Annotator consistency/accuracy with respect to the proxy sentiment classifier for sentiment prediction of translations of original and adversarial examples (adversarial attack is at the input of the NMT system- not the translation). In both settings the proxy classifier is consistent with sentiment perception of human annotations.}
    \label{tab:sent-det}
\end{table}

As the results from the human evaluation experiments correlate to a large extent with the automatic proxy measures/functions in this work, it is appropriate to use the automatic, human-free methods to generate adversarial samples at scale (i.e. using the entire dataset).

\section{Conclusions}

State-of-the-art sequence-to-sequence systems, including Neural Machine Translation (NMT) systems, have demonstrated vulnerability to adversarial attacks. These attacks involve subtle modifications to the input sequence that result in substantial changes in the output sequence, all while being imperceptible to human observers. While existing research has explored adversarial attack methods for NMT systems, focusing on the insertion of target phrases in the output sequences, this study contends that such attacks fail to encompass the entire spectrum of adversarial possibilities.

It is argued that adversaries seeking to manipulate NMT systems may aim to alter the perception of the output translation rather than merely inserting specific phrases. This research sheds light on the ease with which the sentiment perception of NMT system translations can be manipulated. By making minor modifications to the source language text without altering the underlying sentiment, the perception of sentiment in the output translation can be significantly distorted. This finding highlights the importance of considering the susceptibility of NMT systems to perception-based adversarial attacks.

Moving forward, further investigations will be undertaken to explore the robustness of other sequence-to-sequence systems against perception-based adversarial attacks. By broadening the scope of research in this area, a more comprehensive understanding of the vulnerabilities and potential defense mechanisms can be developed, ultimately contributing to the advancement of secure and reliable sequence-to-sequence systems.

\section{Limitations}

This work broadened the concept of adversarial attacks on sequence to sequence systems, where the aim is to change a human's \textit{perception} of the generated output sequence. As a demonstration of this concept, this work explored changing the sentiment perception of the output sequence. It would be useful to also have consideration of other forms of perception of interest, e.g. \textit{fluency} of the generated output sequence for language assessment tasks. This work presents results for four language pairs (de-en, ru-en, en-de, en-ru), and hence the applicability of this work to a language with a vastly different morphology may be limited without explicit experimentation. Finally, the method proposed in this work is applicable to any form of sequence-to-sequence task. Therefore, it would be useful to extend the experiments to tasks beyond machine translation, such as summarisation, question generation, question answering and even grammatical error correction.

\section*{Acknowledgements}

This paper reports on research supported by Cambridge University
Press \& Assessment (CUP\&A), a department of The Chancellor, Masters, and Scholars of the University of Cambridge.

\newpage




\bibliography{main}
\bibliographystyle{icml2021}

\newpage


\end{document}